\documentclass[10pt, a4paper]{article}
\usepackage{lrec}

\usepackage{tikz-dependency}

\usepackage{times}
\usepackage{latexsym}
\usepackage{graphicx}
\usepackage{subcaption}
\usepackage{epstopdf}
\usepackage{tabularx}
\usepackage{soul}
\usepackage{hyperref}
\usepackage{xstring}
\usepackage{xcolor}

\usepackage{url}

\title{An Annotated Dataset of Coreference in English Literature}

\name{David Bamman, Olivia Lewke and Anya Mansoor}

\address{School of Information \\
         University of California, Berkeley\\
        \texttt{\{dbamman,olivialewke,anyamansoor\}@berkeley.edu}\\ }

\abstract{
We present in this work a new dataset of coreference annotations for works of literature in English, covering 29,103 mentions in 210,532 tokens from 100 works of fiction.  This dataset differs from previous coreference datasets in containing documents whose average length (2,105.3 words) is four times longer than other benchmark datasets (463.7 for OntoNotes), and contains examples of difficult coreference problems common in literature.  This dataset allows for an evaluation of cross-domain performance for the task of coreference resolution, and analysis into the characteristics of long-distance within-document coreference.
\\ \newline  }

\begin{document}
\maketitleabstract

\section{Introduction}

Modern coreference resolution systems are typically evaluated on the benchmark OntoNotes dataset\ \cite{hovy_ontonotes:_2006} originally used for the shared tasks on modeling unrestricted coreference at CoNLL 2011\ \cite{pradhan-etal-2011-conll} and CoNLL 2012\ \cite{pradhan-etal-2012-conll}.  Nearly all modern systems evaluate on this data exclusively\ \cite{lee-etal-2017-end,lee-etal-2018-higher,wiseman-etal-2016-learning,clark-manning:2016:P16-1,kantor-globerson-2019-coreference}.

However, the domains included within OntoNotes are relatively narrow---predominantly focused on news (broadcast, magazine, and newswire), conversation, the Bible, and the web---and much recent work has shown the inability of coreference resolution systems to generalize to new domains beyond those they were trained on\ \cite{moosavi-strube-2017-lexical,moosavi-strube-2018-using,subramanian-roth-2019-improving,Lestari2018}.
While coreference datasets exist for other domains such as Wikipedia\ \cite{ghaddar-langlais-2016-wikicoref}, scientific articles\ \cite{schafer-etal-2012-fully} and school examinations\ \cite{chen-etal-2018-preco}, one area lacking robust data is English literature.  Since a growing body of research leveraging computational methods to reason about literature is increasingly relying on coreference information---for example, in the study of literary character \cite{underwood2018,kraicer2018social}---an annotated dataset in this domain is critical for robustly assessing the quality of systems used.

Additionally, as \newcite{roesiger-etal-2018-towards} have outlined, literary texts exhibit markedly different behavior with respect to coreference, rooted in the differing narrative spheres between the world of the narrator and that of the entities in the narration; the level of alternation between generic and specific mentions; the evolution of a character over the space of a long novel; and the differing levels of knowledge that the characters and readers have of the factuality of the events being described.  Each of these phenomena has the potential to impact coreference performance if modeled with a system trained on data where it is rarely seen.

We present in this work a new dataset of coreference annotations for literary texts, to help assess the performance of coreference resolution systems for this domain. 
We make the following contributions:

\begin{itemize}
\item We present a new dataset of 210,532 tokens from 100 different literary novels, annotated for coreference relations between the ACE categories of people, locations, organizations, facilities, geopolitical entities and vehicles\ \cite{ace2005corpus}.
\item We characterize the behavior of coreference phenomena within this dataset, where the average document length is 2,000 words.  This long document length allows us to investigate the temporal distance over which entities exist in discourse, model the burstiness of major entities, and measure the distribution in distance to closest antecedents for different categories of entities; all of these measures can help inform future coreference resolution systems in this domain.
\item We assess the performance of a neural coreference resolution system on this data while varying the domain of the training source.  While a model trained on in-domain literary data achieves an F-score six points higher than one trained on OntoNotes, it performs equivalently to a model trained on PreCo\ \cite{chen-etal-2018-preco}---a very different domain, but two orders of magnitude larger.

\end{itemize}

\section{Related work}

While OntoNotes\ \cite{hovy_ontonotes:_2006} has been the \emph{de facto} benchmark dataset used for coreference resolution, several other resources have been created to explore  coreference in other domains---including Wikipedia articles in WikiCoref\ \cite{ghaddar-langlais-2016-wikicoref}; articles in the ACL Anthology\ \cite{schafer-etal-2012-fully}; school examinations in PreCo\ \cite{chen-etal-2018-preco}; Quiz Bowl questions\ \cite{guha-etal-2015-removing}; and several different aspects of the biomedical domain\ \cite{Cohen2017,nguyen-etal-2011-overview,dsouza,Yang:2004:INP:2105470.2105474,gasperin2007annotation,su2008coreference}.

Less work attends to coreference within specifically literary texts in English, both in terms of annotated data and models for coreference resolution in literature. \newcite{bammanliterary2014} annotates pronominal coreference links within a small sample of five novels for coreference resolution within BookNLP; \ \newcite{vala-etal-2016-annotating} annotates coreference in \emph{Pride and Prejudice}, while \newcite{vala-EtAl:2015:EMNLP} develops a method leveraging character and alias information for character detection in \emph{Sherlock Holmes, Pride and Prejudice} and \emph{The Moonstone}.  The GUM corpus\ \cite{Zeldes2017} contains approximately 12,000 tokens of fiction in its multilayer annotations (which include coreference). \newcite{roesiger-etal-2018-towards} describes a general framework for conceptualizing coreference in literary texts, and presents a tool for annotating coreference in long documents.  Comparatively larger resources, however, exist for German literature---\newcite{krug2017} create a dataset of coreference annotations between character mentions in German novels, which helped assess a rule-based system for coreference resolution \cite{krug-etal-2015-rule}.  

At the same time, coreference in being used as the foundation for empirical arguments in the humanities, including studies that leverage coreference information in order to identify characters and take measurements about them---such as the amount of attention given to characters as a function of their gender\ \cite{underwood2018,kraicer2018social}. 
Several studies have used coreference and alias resolution in the service of identifying character networks\ \cite{Elson2010,agarwal-etal-2012-social,lee-yeung-2012-extracting,Sudhahar2013,jannidis2016,piper2017}.

\section{Data}

We draw our source material for annotation from the texts in LitBank\ \cite{literaryentities}, which consists of 210,532 tokens drawn from 100 different works of English-language fiction from 1719 to 1922 (all within the US public domain).  These texts include a range of literary styles, and include existing annotations for ACE-style entities (people, organizations, locations, geo-political entities, facilities and vehicles), along with \emph{realis} events\ \cite{sims2019}.   
One motivating factor in the corpus selection process was to select a significant sample (approximately 2,000 words) across a wide range of different texts in order to test the performance of systems across different works and authors within the broader domain of literature. 

\section{Annotation}

Our annotation style largely follows that of OntoNotes, in defining the boundaries for markable mentions that can be involved in coreference and in defining the criteria for establishing coreference between them.  However, we make several important departures from their guidelines to accommodate the specific phenomena encountered in literary texts, noted here.

\subsection{Markables}

We make several departures from the OntoNotes definition of a markable span---one that is eligible to be involved in coreference.

\subsubsection{Singletons}

OntoNotes does not allow singleton mentions (noun phrases that are not involved in coreference) to be markable; this decision complicates coreference resolution in test documents, where a separate preprocessing step must be carried out to define the set of candidate noun phrases that are eligible for coreference (i.e., to decide if a noun phrase is a singleton or part a coreference chain).  In the context of OntoNotes, this decision is justified by the presence of a full syntactic parse for all documents; every NP non-terminal in  the parse tree that is not marked as participating in a coreference chain can be inferred to be a singleton.  In our work, we \textbf{do} treat all singleton mentions as markable, removing the need for a gold syntactic parse for each document.   17.4\% of all mentions are singletons in our data, in contrast to the 56\% observed by \newcite{N13-1071} in OntoNotes.

\subsubsection{Entity types}  

While OntoNotes covers unrestricted coreference, we limit the markable entities to only those categories annotated in LitBank.  The existing annotations in Litbank cover six entity types: people (\texttt{PER}), facilities (\texttt{FAC}), locations (\texttt{LOC}), geo-political entities (\texttt{GPE}), organizations (\texttt{ORG}), and vehicles (\texttt{VEH}).  As table \ref{entitytype} shows, the majority (83.1\%) of mentions among these annotations are people, attesting to the significant focus on characters in this domain.

\begin{table}[h!]
\centering
\begin{tabular}{|c|c|c|} \hline
Category&$n$&Frequency \\ \hline\hline
\texttt{PER}&24,180&83.1\% \\ \hline
\texttt{FAC}&2,330&8.0\% \\ \hline
\texttt{LOC}&1,289&4.4\% \\ \hline
\texttt{GPE}&948&3.3\% \\ \hline
\texttt{VEH}&207&0.7\% \\ \hline
\texttt{ORG}&149&0.5\% \\ \hline
\end{tabular}
\caption{\label{entitytype}Counts of entity type.}
\end{table}

\subsubsection{Entity categories}  

Like OntoNotes, we include noun phrases that are proper names (\texttt{PROP}), common phrases (\texttt{NOM}), and pronouns (\texttt{PRON}). The entity annotations in LitBank, however, were originally designed for the task of entity tagging, and focus only on proper nouns (\emph{Tom Sawyer}) and common nouns (\emph{the boy}). 
To correct for this, we include as markable spans all personal pronouns as well, including common forms (\emph{I, me, my, myself, you, your, yourself, she, her, herself, he, him, his, himself, it, its, we, our, they, them, their}), historical forms (\emph{thou, thee, thine, ye}) and forms originating in transcriptions of speech (\emph{'em, 'ee, yeh, yer}).  While annotating a mention, we record its entity category (\texttt{PROP}, \texttt{NOM}, \texttt{PRON}) to enable more fine-grained analysis. As table  \ref{entitycat} illustrates, pronouns account for the majority of mentions in literary texts.

\begin{table}[h!]
\centering
\begin{tabular}{|c|c|c|} \hline
Category&$n$&Frequency \\ \hline\hline
\texttt{PRON}&15,816&54.3\% \\ \hline
\texttt{NOM}&9,737&33.5\% \\ \hline
\texttt{PROP}&3,550&12.2\% \\ \hline
\end{tabular}
\caption{\label{entitycat}Counts of entity category.}
\end{table}

\subsubsection{Quantified and negated noun phrases}

In order to capture potential coreference chains such as ``[No mother]$_x$ should be separated from [her]$_x$ child'', we treat as markables all quantified and negated noun phrases, including those modified by \emph{some, any, many, few, no, neither}, and \emph{none of}.  While OntoNotes rarely annotates these since they would result in singleton coreference chains, we annotate all examples in our dataset.

\subsubsection{Maximal spans}  Following OntoNotes, we mark the maximal extent of a span, as in the following:

\begin{quote}
[The boy who painted the fence and ate lunch] ran away.
\end{quote}

This is notably distinct from the work of \newcite{krug2017}, who only annotate the syntactic heads of characters, and not their full extent.

\subsubsection{Honorifics}

While ACE and LitBank both annotate honorifics in personal names (such as ``[[Mr.] Collins]'', and ``[[Miss] Havisham]]''), these are often stripped by entity recognition systems and coreference resolution systems.  We exclude those from being distinct markable spans here, leaving as entities only the entire maximal span ``[Mr. Collins]'' and ``[Miss Havisham]''.

\subsection{Coreference}

As \newcite{roesiger-etal-2018-towards} point out, there are number of ways in which the style of literary texts influences coreference in a manner different from the domains captured in existing  datasets.  We outline several of those distinctions, along with the decisions we make in annotating them.

\subsubsection{Generic vs. specific mentions}

Most existing datasets for coreference make a distinction between \emph{generic} and \emph{specific} mentions of entities.  A generic noun phrase refers to ``a kind or class of individuals''\ \cite{Reiter:2010:IGN:1858681.1858686}, while a specific mention refers to a specific individual or group.  OntoNotes allows generic mentions to be coreferent only with pronominals and not other generic mentions (``[Doctors]$_x$ care for [their]$_x$ patients);
PreCo additionally allows generic mentions to be coreferent directly (``[Doctors]$_x$ do what is best for [doctors]$_x$'').

We follow the OntoNotes decision here, and disallow generic mentions from being coreferent with each other; importantly, this impacts the annotation of generic ``you'', either used impersonally:

\begin{quote}
Right and left, the streets take [you]$_x$ waterward (Melville, \emph{Moby Dick})
\end{quote}
Or in an address to the reader:
\begin{quote} Let me now take [you]$_x$ on to the day of the assault (Collins, \emph{The Moonstone}).
\end{quote}

One area, however, where the generic/specific distinction becomes difficult is in the use of a nominally generic phrase that may refer ambiguously to a specific individual, as in the following example:

\begin{quote}
Whereas with respect to Turkey, I had much ado to keep him
from being a reproach to me.  His clothes were apt to look oily and
smell of eating-houses.  He wore his pantaloons very loose and baggy in
summer.  His coats were execrable; his hat not to be handled.  But while
the hat was a thing of indifference to me, inasmuch as his natural
civility and deference, as a dependent Englishman, always led him to
doff it the moment he entered the room, yet his coat was another matter.
Concerning his coats, I reasoned with him; but with no effect.  The
truth was, I suppose, that [a man of so small an income] could not afford
to sport such a lustrous face and a lustrous coat at one and the same
time. (Melville, \emph{Bartleby, The Scrivener})
\end{quote}

Here Melville describes the physical appearance of the character Turkey in \emph{Bartleby, The Scrivener}; while the generic entity ``a man of so small an income'' surely is being used to describe Turkey, it refers to a more general class of entity, of which Turkey is only a member.

We can see this to be an example of \emph{class near-identity}\ \cite{recasens-etal-2010-typology}, where ``[Turkey]'' and ``[a man of so small an income]'' share an \emph{is-a} relation, with latter being a more general realization of the former. In these cases, while the generic mention may be interpreted as referring to a specific individual, we still treat it as subject to the guidelines for generics, and disallow coreference between the generic mention and any specific individual.

\subsubsection{Copulae}

Existing datasets like OntoNotes and PreCo differ in their annotation of copulae---structures where a given entity is linked to an attribute through a copular verb such as \emph{be, appear, feel, seem}, etc, as in {``[John] is [a doctor]''}.  Datasets either rule out their eligibility for coreference at all (as in OntoNotes) or as being fully coreferent with an entire coreference chain (PreCo).  While PreCo allows copular attributes to be coreferent with their subject, we follow OntoNotes in seeing the attributes in copular structures as not fundamentally referential in nature; it is the act of predication that associates two mentions, not their referentiality.  We can see this in part by examining the range of copular structures present in our literary dataset, in which we can clearly see that the strength of the association between a subject and attribute is moderated by the form of the linking verb (and any adverbials that modify it).  While example \ref{easy} below is a copular structure that clearly asserts the equivalence of the comparands, \ref{inmanyways} is not---``in many ways'' suggests that the predication is not complete (there exist some ways in which he was not ``a most valuable person'').  Examples \ref{past1} and \ref{past2} are copular structures about assertions in the past, which may no longer hold in the present; and examples \ref{neg1} and \ref{neg2} are both copular but negated.

\begin{enumerate}
    \item \label{easy} {[I]} am [a rather elderly man] (Melville, \emph{Bartleby}).
\item \label{inmanyways} {[He]} was {in many ways} [a most valuable person to me].  (Melville, \emph{Bartleby}).
\item \label{past1} Mrs. Marjoribanks, poor lady, {had been} an invalid for many years (Oliphant, \emph{Miss Marjoribanks}).
\item \label{past2} {[It]} was [a convent] before the Thirty Year's War (Von Arnim, \emph{Elizabeth and Her German Garden}).
\item \label{neg1} {[She]} was \emph{not} to be described as [a tall girl] (Oliphant, \emph{Miss Marjoribanks}).
\item \label{neg2} {[Lucilla]} was \emph{not} [the woman to be disconcerted] (Oliphant, \emph{Miss Marjoribanks}).

\end{enumerate}

The ways in which a copula can serve as an identity-linking function between two mentions is clearly on a continuum.  To capture just those that are identity-establishing, we only annotate examples like \ref{easy} above: only those that are asserted as currently holding true, not relations that might hold in the future, hypotheticals, or negation.

And since it is the copular structure itself that asserts identity rather than the referentiality of the mention, we annotate copulae as a link between individual mentions (as in the OntoNotes treatment of appositions), so that example (1) would be annotated as the following:

\begin{center}
\begin{dependency}[theme = default,
label style={font=\bfseries,thick}
]
   \begin{deptext}[column sep=-.3em, row sep=.0ex]

\textrm{[I]$_x$}\&am$\;$\& \textrm{[a rather elderly man]}\\
   \end{deptext}
   \depedge[edge unit distance=1.5ex]{3}{1}{cop}
\end{dependency}
\end{center}

\subsubsection{Apposition}

Like OntoNotes, we also annotate apposition as a distinct relation that holds between specific mentions that are immediately adjacent:
\begin{center}
\begin{dependency}[theme = default,
label style={font=\bfseries,thick}
]
   \begin{deptext}[column sep=-.3em, row sep=.0ex]

\textrm{[The Nellie]$_x$}\&,$\;$\& \textrm{[a cruising yawl]}\&,$\;$\& swung \& to \& \textrm{[her]$_x$} \&anchor\\
   \end{deptext}
   \depedge[edge unit distance=1.5ex]{3}{1}{appos}
\end{dependency}
\end{center}

\subsubsection{Identity and Near-Identity}

Classical models of coreference often characterize the problem as deciding whether two mentions refer to the same entity in the real world, which can easily become entangled in  metaphysical complexities on the nature of identity (for an overview, see \newcite{gallois2016}.  While all discourse involves such complexities, it is exacerbated in literary novels, which not only describe entities that may exist, and change, over the course of a long narrative timeline, but may also describe the historical background for those entities.  Novels may describe a city as it evolves over the course of a millennium (such as London), or the development of a child into an adult over a period of decades (\emph{Great Expectations}).  Does the London of 1922 refer to the same entity as London in 1599? And does the seven-year-old Pip at the beginning of \emph{Great Expectations} refer to the same entity as the thirty-year-old Pip at the end?

Rather than determine the identity of reference of real-world entities, we draw on the formalization of \newcite{RECASENS20111138} in their discussion of near-identity in coreference, in which they outline the specific operations of {neutralization}/compression and {refocusing}/decompression that respectively minimize or maximize the differences in feature values between two discourse entities, effectively making them \emph{more} identical in the case of neutralization and \emph{less} identical in the case of refocusing.  Importantly, the degree of identity in coreference here is tied not to entities that exist in the real world, but rather to discourse entities constructed in a discourse, and for whom a specific pragmatic context may encourage two mentions to be seen as more or less coherent---independent of any real-world status they might have\ \cite{nunberg1984}.
 \newcite{RECASENS20111138}  cite the example of \emph{Postville} to illustrate pragmatic near-identity:

\begin{quote}
On homecoming night [Postville]$_x$ feels like Hometown, USA, but a look around [this town of 2,000]$_x$ shows it's become a miniature Ellis Island. This was an all-white, all-Christian community . . . For those who prefer [the old Postville]$_y$, Mayor John Hyman has a simple answer. \cite[10]{RECASENS20111138} 

\end{quote}

Here \emph{Postville} and \emph{the old Postville} are pragmatically distanced through an act of refocusing to emphasize their differences (the inhabitants of old Postville being predominantly white, while the Postville contemporaneous with writing is more diverse).

In Hawthorne's \emph{House of the Seven Gables} we see a similar case of near-identity involving entities that exist with the same physical boundaries but different temporal extent; here, however, we see a reverse effect of compression, flattening the temporal differences between the eponymous \emph{House of the Seven Gables} and \emph{the old Pyncheon House}:

\begin{quote}
Halfway down a by-street of one of our New England towns stands [a rusty
wooden house, with seven acutely peaked gables, facing towards various
points of the compass, and a huge, clustered chimney in the midst]$_{x}$.  The
street is Pyncheon Street; [the house]$_{x}$ is [the old Pyncheon House]$_{cop\rightarrow {{the\; house}}}$.

\end{quote}

Here, even though the houses have different feature values (different names and different temporal extents), compression equates them through the use of copular predication.
At the same time, this entity is distanced through refocusing with respect to an temporally earlier entity that occupied the same physical location:

\begin{quote}
[The House of the Seven Gables]$_{x}$, antique as it now looks, was not [the
first habitation erected by civilized man on precisely the same spot of
ground].  \ldots A natural spring of soft
and pleasant water \ldots had early induced Matthew Maule to build [a
hut, shaggy with thatch, at this point, although somewhat too remote
from what was then the centre of the village]$_{y}$.
\end{quote}

We can see another example of this compression in Orczy's \emph{The Scarlet Pimpernel}:

\begin{quote}
During the greater part of the day the guillotine had been kept busy at
its ghastly work: all that [France]$_x$ had boasted of in the past centuries,
of ancient names, and blue blood, had paid toll to [her]$_x$ desire for
liberty and for fraternity. 
\end{quote}

\emph{The Scarlet Pimpernel} is set in France during the French Revolution, just after the regime change the revolution entailed.  Under a metaphysical criterion of identity, we might distinguish between  France as a monarchy prior to revolution and France as a republic under the Committee of Public Safety (assuming we think of a state as being comprised not only of its territorial holdings, but the structure of its government and the customs/practices of its people).  Here, however, Orczy neutralizes those shades of meaning by stressing the France that has persisted through centuries; we treat all mentions of France as coreferent under this compression.

\subsubsection{Revelation of identity}

One other phenomenon that is more characteristic of literary texts than texts from other domains is a revelation of identity, when one entity is revealed to be identical with another.  This revelation can be sudden, in which entities $x$ and $y$ were depicted as being separate but are revealed to be the same, as in the case of \emph{whodunnit} detective novels and mysteries. In \emph{Great Expectations}, Pip encounters a convict as a child at the beginning of the novel, and is later supported as a young man by an unknown benefactor; these are revealed to be the same entity at the end of the novel.  Here we adopt an approach similar to \newcite{roesiger-etal-2018-towards}, which annotates identity from the reader's point of view (as distinct from characters'); here, we can see that the convict and Pip's benefactor are identical if we presume that a reader's point of view is scoped over the entirety of the text.

Revelations of identity can also arise through a gradual accumulation of information. In Conan Doyle's \emph{Adventures of Sherlock Holmes}, Holmes receives a letter noting the imminent arrival of a visitor:
\begin{quote}
``There will call upon you to-night, at a quarter to eight o'clock,'' it said, ``[a gentleman who desires to consult you upon a matter of the very deepest moment]$_x$. 
\ldots Be in your chamber then at that hour, and do not take it amiss if [your visitor]$_x$ wear a mask.''
\end{quote}

Here the fact that the identity of the visitor described in the letter is the same as that of the author of the letter is not known to the reader; Sherlock then proceeds to deduce that the author of the letter is likely a German man and is the same individual described as the imminent visitor, concluding:

\begin{quote}
``It only remains, therefore, to discover what is wanted by [this German who writes upon Bohemian paper and prefers wearing a mask to showing [his]$_x$ face]$_x$. And here [he]$_x$ comes, if I am not mistaken, to resolve all our doubts.''
\end{quote}

The equivalence of these entities is pragmatically determined only after several mentions (and realized syntactically through a conjunction).  While a reader could not know that the visitor is the same individual as the author when reading the letter above, we treat all mentions as coreferent if a reader can determine that they are identical at any point in the narrative.  In other words, we annotate from the perspective of the textual reality, rather than from the reader's state of knowledge at the moment of encountering the mention to be resolved.

\subsection{Annotation process}

We carry out annotations in a two-step process: first using a custom command-line interface for linking mentions to entities in text\footnote{\url{https://github.com/dbamman/cl-coref-annotator}} and then transferring the annotations to a BRAT\ \cite{Stenetorp:2012:BWT:2380921.2380942} GUI interface for checking.

All annotations were performed by three annotators (all three authors) after an initial phase of  guideline design and consistency checking.  While each text was annotated by a single author, we double-annotated a total of 10 full texts (10\% of the total collection) in order to assess the consistency of annotation across different annotators.  As with prior work, we calculate inter-annotator agreement using the same coreference metrics used for assessing the performance of coreference resolution systems; while OntoNotes reports an average inter-annotator MUC score of 83.0 across its seven subdomains\ \cite{pradhan-etal-2012-conll} and PreCo reports a score of 77.5, we find very high agreement between our trained annotators with a MUC score of 95.5.\footnote{The inter-annotator agreement for all three commonly used coreference metrics are: 93.81 $B^3$,  95.53 MUC, and 87.72 CEAF$_{\phi_4}$.}  We suspect this is due to a combination of two factors: the smaller pool of high-trained annotators we use for our smaller collection, and the restricted nature of the entities we are labeling (only people, facilities, locations, geopolitical entities, organizations and vehicles) instead of unrestricted coreference.  This data is publicly available as a part of LitBank at \url{https://github.com/dbamman/litbank}.

\section{Analysis}\label{analysis}

We carry out several analyses to investigate the behavior of coreference within this dataset.

\subsection{Spread}

\begin{figure}[h]
\begin{centering}
\includegraphics[scale=.7]{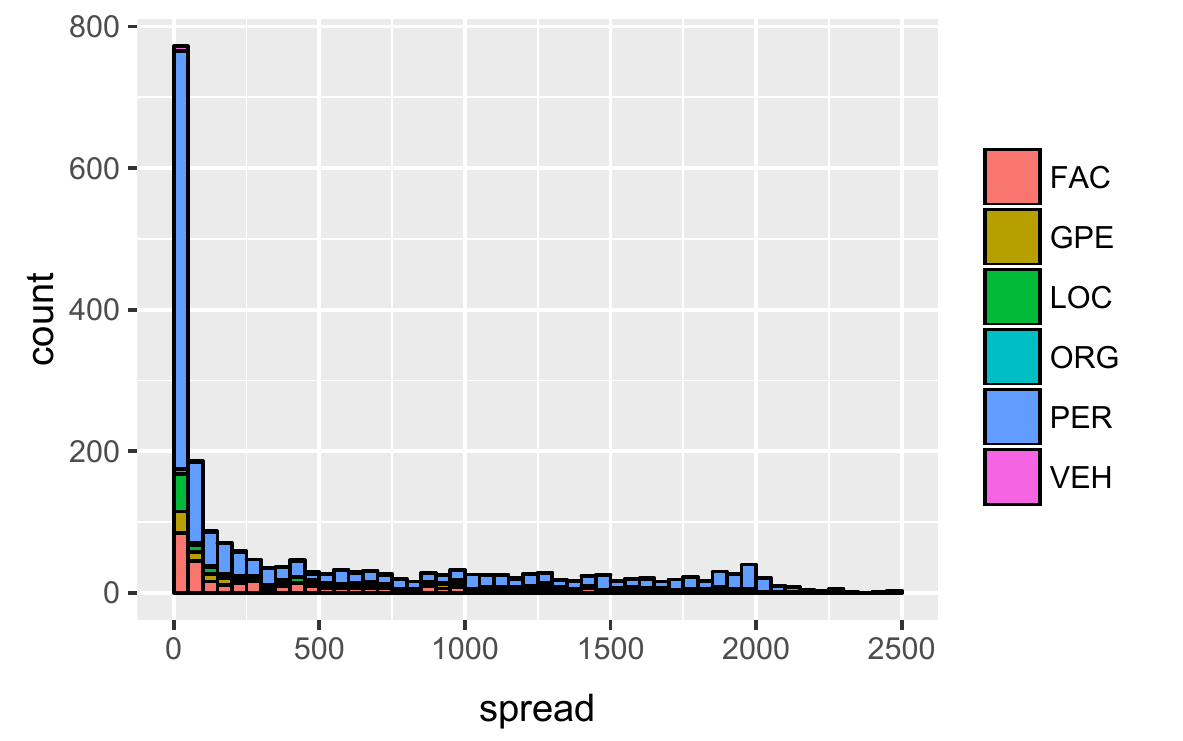}
\caption{\label{spreadent} Spread between first and last mention of entity, in tokens.}
\end{centering}
\end{figure}

Figure \ref{spreadent} examines the distribution in the temporal distance (in narrative time) in which a given entity is active, where distance is measured as the number of tokens between the first mention of an entity and the last mention. The vast majority of entities span relatively short time scales; 50\% of them span 173 or fewer tokens (approximately half a page in a printed book).  These entities include both generic mentions (\emph{a warrior}), known entities grounded in the real world (\emph{New York}) and specific entities mentioned only briefly that do not factor prominently in the broader discourse.

\begin{figure}[h]
\begin{centering}
\includegraphics[scale=.7]{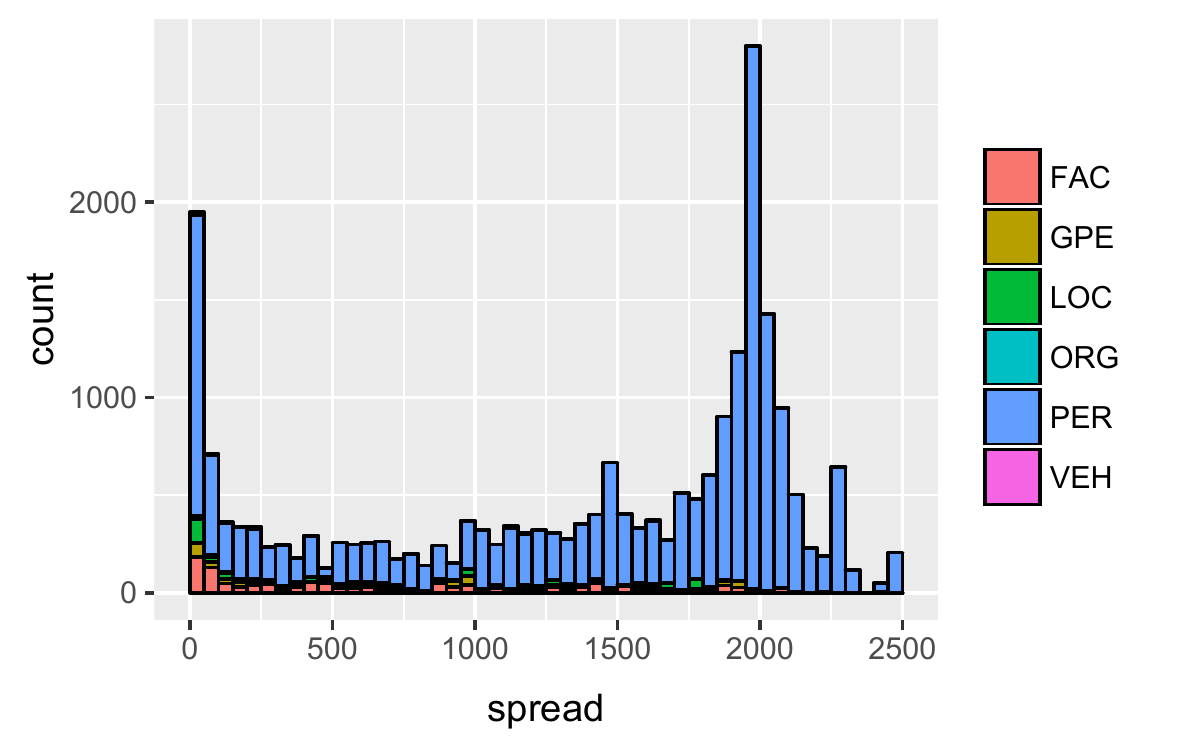}
\caption{\label{spreadtok} Spread between first and last mention of entity, in tokens, weighted by the number of mentions in each entity.}
\end{centering}
\end{figure}

In figure  \ref{spreadent}, each entity is counted equally as a single data point, giving equal weight to major characters and minor characters. Figure \ref{spreadtok}  in contrast examines the same phenomenon from the perspective of individual mentions, where each entity is weighted by the number of mentions associated with it.  As can be seen, while most entity \emph{types} span relatively short distances, most entity \emph{tokens} are part of coreference chains that span the entire document (ca. 2,000 words).  Major characters, in particular, constitute the majority of coreferential mentions.

\subsection{Burstiness}

To explore whether entities that span long text ranges tend to cluster together in tight bursts, we quantify the burstiness of an entity by measuring the entropy of the empirical distribution defined over mentions in narrative time: we divide the narrative time of a book into 100 equal-sized segments (each spanning roughly 200 words), and define $d_{e,i}$ to be the relative frequency with which entity $e$ was mentioned in segment $i$ (relative to the total number of mentions of $e$ in the document).  We calculate entropy as $H(d_e)$.

\begin{figure}[h]
\begin{centering}
\includegraphics[scale=.63]{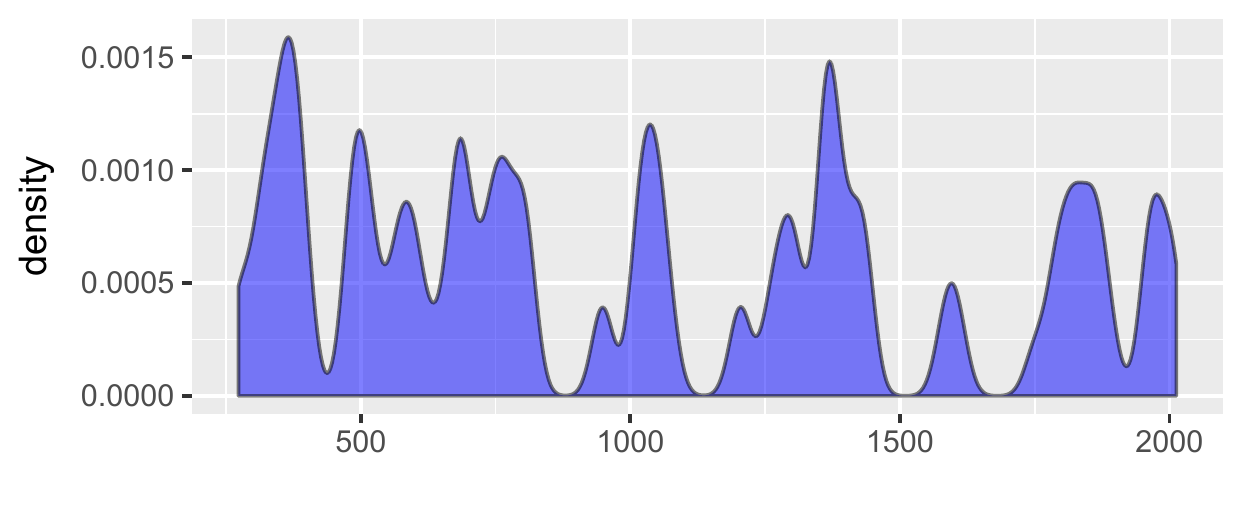}
\includegraphics[scale=.63]{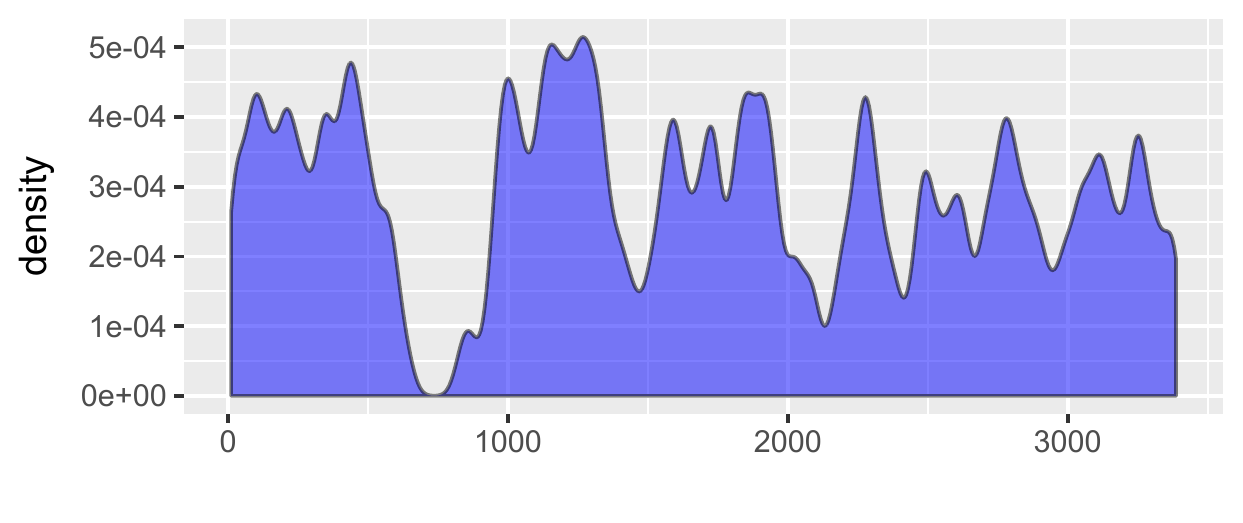}
\caption{\label{bursts} Long-range entities are bursty; the distribution of mentions over narrative time for the entity with the lowest entropy (top; Basil Hallward in Wilde's \emph{The Picture of Dorian Gray}) and highest entropy (bottom; the narrator in Swift's \emph{Gulliver's Travels}).}
\end{centering}
\end{figure}

Figure \ref{bursts} illustrates this bursty behavior visually by selecting the entities with the lowest entropy (Basil Hallward in \emph{The Picture of Dorian Gray}) and highest entropy (the narrator in \emph{Gulliver's Travels}) among those spanning at least 1500 tokens and mentioned at least 100 times.  While Basil Hallward goes through several periods of not being mentioned followed by increased focus (bursty behavior), even the narrator of \emph{Gulliver's Travels} exhibits bursty behavior despite being more uniformly mentioned, with a gap of several hundred words in which he is not mentioned between his otherwise constant focus.  Even entities that have high entropy (which should be closer to a uniform distribution of mentions over time) still exhibit bursty behavior in which there is a period of time where they are not mentioned.

\subsection{Distance to nearest antecedent}

\begin{figure}[h]
\begin{centering}
\includegraphics[scale=1]{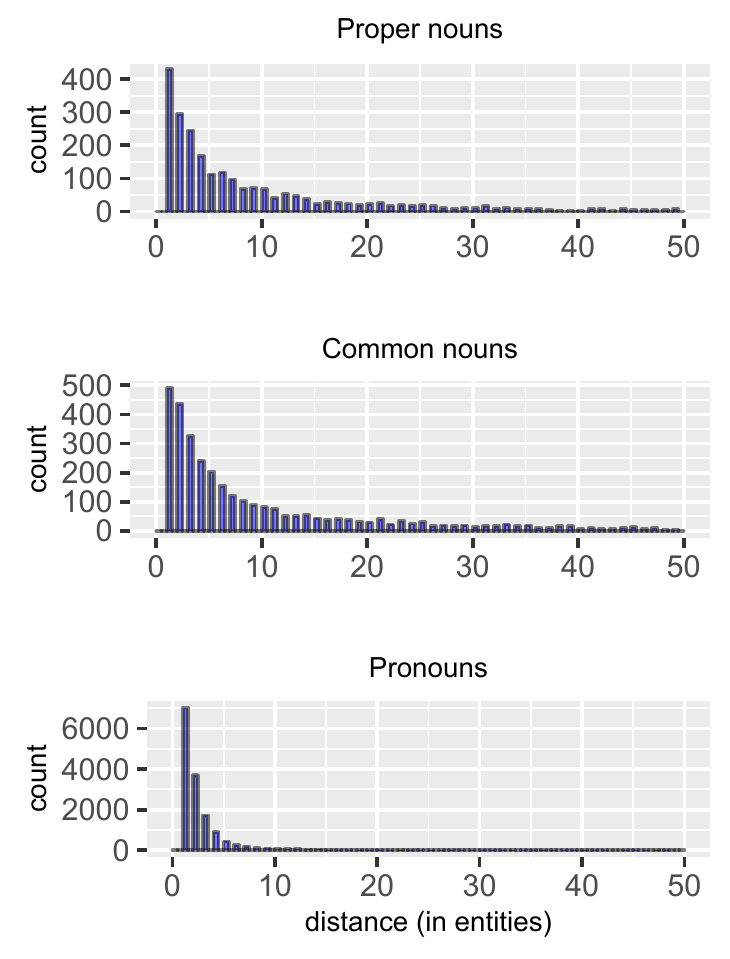}
\caption{\label{distances} Distance to antecedent in entities.}
\end{centering}
\end{figure}

Finally, we examine the distribution in distances to the closest antecedent for proper nouns, common nouns and pronouns for non-singleton mentions, as depicted in figure \ref{distances}.  Proper nouns  have a median distance of 5 entities to their nearest antecedent, with 90\% of antecedents appearing within 42 mentions, and 95\% of antecedents within 78.3 mentions.  Common nouns (3572 mentions) have a median distance of 6 entities to their nearest antecedent, with 90\% of antecedents appearing within 59 mentions, and 95\% of antecedents within 101 mentions. Pronouns have a median distance of 2 entities to their nearest antecedent, with 90\% of antecedents appearing within 5 mentions, and 95\% of antecedents within 9 mentions.   While coreference systems often impose strict limits to the number of maximum antecedents to consider for long documents\ \cite{lee-etal-2017-end}, or use coarse-to-fine inference for reducing the number of candidates\ \cite{lee-etal-2018-higher}, this suggests that pronouns (which again account for over half of all potentially coreferential mentions in this data) only need to consider a far shorter number of antecedents.

\section{Empirical performance}

The existence of a dataset of coreference in works of English literature allows us to test the performance of coreference resolution systems in this domain---including those trained on other benchmark datasets (OntoNotes and PreCo) and those trained on our newly created  coreference annotations in LitBank.

\subsection{Gold mentions}

OntoNotes, PreCo and LitBank all differ in their annotations in two important ways: not only in the individual coreference decisions (where, for example, OntoNotes and LitBank do \emph{not} link two generic common nouns, while PreCo does), but also in the core annotation of mentions---while OntoNotes does not annotate singleton mentions (those entities that are not coreferent with any other), PreCo and LitBank both do.  Additionally, while OntoNotes and PreCo both capture unrestricted entity coreference (annotating all entities regardless of their entity type), LitBank only annotates those that fall in the ACE entity categories \texttt{PER},  \texttt{FAC},  \texttt{LOC},  \texttt{GPE},  \texttt{ORG}, and  \texttt{VEH}.  In order to assess the potential for models trained on these datasets to accurately capture this kind of coreference, we first evaluate on gold mention spans in LitBank.

For each dataset, we train a neural model based on \newcite{lee-etal-2017-end}.  While \newcite{lee-etal-2017-end} is an end-to-end mention-ranking model that jointly performs mention identification and linking, we constrain it to fixed mention spans and use it solely for mention ranking, greedily selecting the most likely antecedent for each mention in a document (or a null antecedent that begins a new coreference chain with that mention).  We also differ by representing each token in a sentence not by static Glove/Turian word embeddings and character convolution, but through its BERT token representation\ \cite{devlin-etal-2019-bert}.  While BERT uses WordPiece tokenization\ \cite{DBLP:journals/corr/WuSCLNMKCGMKSJL16}, we average together the individual WordPiece tokens for a given word to form its token representation, as in \newcite{sims2019}.

This model is a bidirectional LSTM that generates  output $x_i$ for each token at position $i$.  A mention $m$ spanning sentence positions $[start, end]$ is represented as the concatenation of $x_{start}$, $x_{end}$, the output of a learned attention mechanism over [$x_{start}, \ldots, x_{end}$], and features expressing the width of the mention span and whether the span falls within a quotation (each feature is embedded in its own representation space).  Given two mention representations $g_i$ and $g_j$, the score of their linking is a feedforward network over the concatenation of $g_i$, $g_j$, the elementwise product of $g_i$ and $g_j$, and a feature function scoped over the two mentions; we use features expressing the distance between mentions (in mentions and sentences) and whether one mention is nested within the other.  This model is trained to maximize the marginal log-likelihood of all antecedents in the correct coreference chain for each mention; during prediction, we proceed from the beginning of the document to the end, and greedily select the single highest-scoring antecedent (or the null antecedent) for each mention, and define a coreference chain as the transitive closure of all such links. 

To summarize, this model is identical with  \newcite{lee-etal-2017-end} with the following exceptions: 1.) We use BERT contextual representations instead of static word vectors and a subword character CNN (to make use of advancements in representation learning); 2.) we train and predict conditioning on mention boundaries (in order to separate the core task of mention linking from mention identification, given the differences in mention boundaries in the three datasets; 3.) We omit author and genre information from training (as these are relevant only in OntoNotes and not in PreCo or LitBank); 4.) We only consider antecedents within 20 mentions for pronouns and 300 mentions for proper noun phrases and common noun phrases (given the observations on distance distributions above).

Aside from these differences, we preserve the same core hyperparameter choices in \newcite{lee-etal-2017-end}: LSTM size of 200; the feedforward network is comprised of two 150-dimensional layers; each feature is embedded in a learned 20-dimensional space; and we use ADAM\ \cite{Adam} for learning, decaying the learning rate by 0.1\% each 100 steps.  Each training source has a training and development split; we train each model until there is no improvement for 10 epochs on its development set, saving the best-performing model on that development data.  We test performance on literary data by training on three different datasets: OntoNotes, PreCo and our literary data; we evaluate on the entirety of the literary annotations (using cross-validation where needed, as described below), excluding copula and apposition links. Table \ref{goldperf} presents the results of this evaluation using F-scores from $B^3$\ \cite{bagga1998algorithms}, MUC\ \cite{Vilain:1995:MCS:1072399.1072405} and CEAF$_{\phi_4}$\ \cite{Luo:2005:CRP:1220575.1220579}, along with their average.

\paragraph{OntoNotes.}
The training set of OntoNotes contains a total of 1.3M tokens; when evaluated on its own test set, we see performance comparable to that reported in \newcite{lee-etal-2017-end} for gold mentions, along with an expected drop in performance for not using speaker or genre metadata (83.2 vs. 85.2).  When evaluated on LitBank, however, this performance drops 10.3  points to 72.9.

\paragraph{PreCo.}
The training set of PreCo contains an order of magnitude more data, with a total of 12.2M tokens.
This increase in training size relative to OntoNotes translates into an improvement in overall accuracy: when evaluated on LitBank, we see an average F1 score of 78.8.

\paragraph{LitBank.}
To evaluate on LitBank, we perform a 10-fold cross-validation---training a model on 80\% of the data, assessing early stopping using a development set of 10\% of the data, and then evaluating that trained model on the remaining held-out 10\%---over all ten partitions.  Such a model achieves an average F-score of 79.3, indistinguishable from a model trained on PreCo but substantially better than one trained on OntoNotes.

\begin{table}[h!]
\centering
\begin{tabular}{|c|c|c|c|c|} \hline
Training source&$B^3$&MUC&CEAF$_{\phi_4}$&Average \\ \hline \hline
OntoNotes&66.9&85.7&65.9&72.9 \\ \hline
PreCo&73.8&88.4&74.3&78.8\\ \hline
LitBank&72.6&88.5&76.7&79.3\\ \hline
\end{tabular}
\caption{\label{goldperf}Coreference resolution performance on gold mentions.}
\end{table}

A natural reason why OntoNotes may underperform on this dataset is the mismatch between annotation guidelines---especially the difference between the choice by PreCo and LitBank to annotate singleton mentions and OntoNotes not to.  To assess this possibility, we can leverage the $B^3$ coreference metric to interrogate performance on a subset of the data. Precision (P) and recall (R) in $B^3$ is calculated by finding the overlap between a gold coreference chain $\mathcal{G}_i$ and system coreference chain $\mathcal{S}_i$ associated with each mention $i$ over all $N$ mentions:
$$
\textrm{P} = {1 \over N} \sum_{i=1}^N {{|\mathcal{G}_i \cap \mathcal{S}_i|} \over |\mathcal{S}_i|}\quad\quad\quad
\textrm{R} = {1 \over N} \sum_{i=1}^N {{|\mathcal{G}_i \cap \mathcal{S}_i|} \over |\mathcal{G}_i|}
$$

We can adapt this to examine non-singleton performance by ranging only over mentions that are not singletons in the gold data; if we exclude true singleton mentions from our evaluation, the performance for all three training sources drops (since singletons are comparatively easier to identify)---with gold mention boundaries, PreCo $B^3$ drops from 73.8 to 69.7, OntoNotes $B^3$ drops from 66.9 to 63.3, and LitBank $B^3$ drops from 72.6 to 67.7.  OntoNotes drops less than the other methods---attesting to some impact of the singleton annotation difference---but still lags behind the other two sources of data on this metric.

\subsection{Predicted mentions}

To test real-world performance when not given access to gold mention boundaries, we build a mention identifier trained on the mention boundaries in our annotated literary data, and assess coreference resolution performance when using these predicted mentions at test time.

\subsubsection{Mention identification}

We train a mention identification system with a layered BiLSTM-CRF of \newcite{ju-etal-2018-neural} used in \newcite{literaryentities}; we adapt it here by replacing the static word embeddings in the original work with BERT contextual embeddings.  In a tenfold cross-validation on our annotated data to predict the mention spans, we find an F-score for mention detection to be 89.1.  Adding a prediction for whether an entity is a proper noun phrase, common noun phrase or pronoun only decreases the F-score to 88.3; adding a prediction for the entity class (person, organization, location, facility, geo-political entity, or vehicle)  yields a performance of 87.3.

\begin{table}[h!]
\centering
\begin{tabular}{|l|c|c|c|} \hline
Task&Precision&Recall&F\\ \hline \hline
Mention span detection&90.7&87.6&89.1 \\ \hline
\quad + \texttt{PROP}/\texttt{NOM}/\texttt{PRON}&90.2&86.5&88.3 \\ \hline
\quad + Entity class&89.2&85.5&87.3 \\ \hline
\end{tabular}
\caption{\label{mentions}Mention identification performance.}
\end{table}

\subsubsection{Coreference resolution performance}

To assess performance on predicted mentions, we carry out another ten-fold cross-validation: for one partition of the data into 80\% training data, 10\% development data, and 10\% test data, we use a mention identification model trained on the training split to predict mention boundaries in the test split, and then train our neural coreference system on that same train split to make coreference decisions on the predicted mentions in the test split (again using the development set to assess early stopping).  Table \ref{predicted_mentions} presents the results of this evaluation.  While performance naturally degrades as a function of the predicted mentions, we see the same overall rank between training sources as found in table \ref{goldperf}: LitBank (68.1F) and PreCo (67.6F) are indistinguishable but both perform substantially better than OntoNotes.

\begin{table}[h!]
\centering
\begin{tabular}{|c|c|c|c|c|} \hline
Training source&$B^3$&MUC&CEAF$_{\phi_4}$&Average \\ \hline \hline
OntoNotes&57.7&81.2&49.7&62.9 \\ \hline
PreCo&63.5&84.2&55.1&67.6\\ \hline
LitBank&62.7&84.3&57.3&68.1\\ \hline
\end{tabular}
\caption{\label{predicted_mentions}Coreference resolution performance on predicted mentions.}
\end{table}

\section{Conclusion}

We present in this work a new dataset of coreference annotations for 210,532 tokens drawn from 100 different literary texts, to allow the analysis of coreference performance in a wide range of literary styles and authors, and to train coreference resolution systems on literary data in English.  As more and more work in literary analysis makes use of computational methods to explore the persistence of entities in text---from characters\ \cite{piper2018,underwood2019} to objects\ \cite{tenen2018}---having reliable estimates of the quality of different coreference methods is critical for assessing their impact on downstream measurements.

In addition to the core contribution of this dataset and evaluation, there are three takeaways worth highlighting: the first is the relatively high performance of PreCo in cross-domain coreference resolution.  While PreCo comes from a separate domain of school examinations, it is able to match the performance of a smaller dataset of in-domain coreference annotations; while the literary annotations are able to achieve comparable performance with two orders of magnitude less data, PreCo illustrates the power of large annotated datasets to work well across a potential variety of domains.  Second, we illustrate the behavior of entities across long spans of text---demonstrating that entities tend to be bimodal (either active over very short timespans or very long ones), and frequent entities tend to exhibit bursty behavior over the course of long documents.  Third, we illustrate the differing antecedent behavior of different mention types, showing that pronouns in particular can be linked to an antecedent within nine mentions 95\% of the time.  We hope this work can spur future work on literary coreference, since the characteristics of literature---the relatively long length of documents, the centrality of relatively few major entities, the difficult  metaphysical and epistemological questions of identity and the revelation of  knowledge---can potentially provide a unique vantage point on the problem of coreference in general.  This data is freely available at \url{https://github.com/dbamman/litbank}; code to support this work can be found at \url{https://github.com/dbamman/lrec2020-coref}.

\section{Acknowledgments}

The research reported in this article was supported by an Amazon Research Award and by resources provided by NVIDIA and Berkeley Research Computing.

\section{Bibliographical references}

\bibliography{lrec}
\bibliographystyle{lrec}

\end{document}